\documentclass[letterpaper, 10 pt, conference]{ieeeconf}  

\IEEEoverridecommandlockouts                              

\usepackage[utf8]{inputenc} 
\usepackage[T1]{fontenc}    
\usepackage{url}            
\usepackage{booktabs}       
\usepackage{amsfonts}       
\usepackage{xfrac}
\usepackage{microtype}      
\usepackage{xcolor}         
\usepackage{colortbl}       

\usepackage{listings}
\usepackage{soul}
\usepackage{lipsum}

\usepackage{graphicx, subfigure}
\usepackage{multirow, longtable, diagbox, setspace, rotating}
\usepackage{wrapfig, float, caption}

\usepackage{amsmath, amssymb, amsfonts} 
\usepackage{mathtools}
\usepackage[ruled, vlined]{algorithm2e}
\usepackage{algpseudocode}

\usepackage{tikz}
\usetikzlibrary{arrows.meta, positioning, shapes.geometric, calc, decorations.pathmorphing}
\usepackage[hidelinks]{hyperref}

\title{\LARGE \bf
Automatic Labelling for Bimanual Mobile Manipulation
}

\author{
Yupu Lu$^{1}$ and Jia Pan$^{1}$%
\thanks{$^{1}$School of Computing and Data Science, The University of Hong Kong, HKSAR. 
        {\tt\small Correspondence: \{luyp16,panj\}@connect.hku.hk}}%
}

\begin{document}
\maketitle
\thispagestyle{empty}
\pagestyle{empty}

\begin{abstract}
Semantically meaningful subtask labels can provide useful contexts for long-horizon policies, but automatically identifying both reliable temporal boundaries and broad semantic descriptions for annotations remains difficult.  We present an automatic labelling pipeline that assigns temporal localisation to deterministic trajectory analysis and semantic interpretation to vision-language (VL) reasoning.  The pipeline segments synchronised kinematic signals into phases, performs phase-localised VL reasoning to describe the contents, and aggregates the outputs for the base, left arm, and right arm actions.  
We evaluate this pipeline primarily on 29 real Galaxea bimanual mobile-manipulation tasks.  Repeating the VL reasoning three times first produces the same output value for 87.4\% on selected tasks.  A review by nine participants across all 29 tasks then judgements on the labelled phases and shows positive acceptance of temporal divisions (90.5\%), body labels (90.7\%), and arm labels (78.7\%).  The results indicate that the segmentation--VL design can produce structured annotations while preserving asynchronous bimanual behaviour, providing a basis for richer semantic subtask identification and state-based verification.
\end{abstract}


\section{INTRODUCTION}
\label{sec:intro}

Long-horizon manipulation demonstrations contain concurrent motions whose temporal organisation and semantic interpretation are both relevant to learning and analysis.  Dense subtask annotations can expose what happens, which physical stream is involved, and when it occurs. These are benefitial for those policy learning methods like hierarchical diffusion policies~\cite{ma2024hierarchical}, vision-language-action models~\cite{wu2026pragmatic} to align trajectories to subtask skills. But such annotations usually require manual labelling, which is time and resource consuming.

To obtain annotations automatically, the labelling system generally relies on either kinematic heuristics or direct vision-language (VL) inference.  Kinematic data provide precise and inspectable temporal cues, but the applicability and semantic coverage of heuristic methods are task specific~\cite{yu2024bikc}.  VL reasoning covers a broader range of actions from visual observations, yet hard to determine the exact timing of events~\cite{lu2026keypose}. 

We address their limitations by combining the strengths of both approaches.  Deterministic trajectory analysis first establishes a temporal division from the recorded kinematics; phase-localised VL reasoning then interprets the divided phases without estimating their boundaries; and deterministic aggregation preserves the base, left-arm, and right-arm streams.  In this way, the pipeline combines the temporal precision of heuristic segmentation with the semantic breadth of VL reasoning, producing records on both when an interaction occurs and what is visible in that phase.

Our contributions are:
\begin{itemize}
  \item an automatic construction path that combines deterministic trajectory-phase generation, phase-localised VL description, and independent semantic-stream aggregation;
  \item a multi-signal trajectory segmenter over per-arm motion, mobile-base motion, torso activity, and command-level gripper events, with explicit cross-stream synchronisation;
  \item a per-phase VL reasoner, followed by a logical row-first semantic aggregation to combine adjacent phases with compatible descriptions;
  \item a Galaxea dataset evaluation of reproducibility and participant acceptability for the resulting temporal divisions and row-level semantic streams, together with extra representative task analyses.
\end{itemize}

\section{RELATED WORK}
\label{sec:related}

\subsection{Temporal and Semantic Supervision for Policies}
To represent different skills or task stages, keypose-conditioned policies use selected robot configurations as intermediate targets, separating the choice of an interaction goal from the generation of the motion that reaches it~\cite{ma2024hierarchical,yu2024bikc,xu2025bikc+}.  Subtask descriptions provide a complementary form of conditioning in recent vision-language-action systems~\cite{intelligence2026pi,wu2026pragmatic}, expressing the intended activity in language rather than configuration space.  These policies motivate for annotations that associate action meaning with specific portions of a demonstration.  The question considered here is how to better obtain such temporal and semantic information automatically from these recorded execution.

\subsection{Manipulation Datasets and the Annotation Bottleneck}
Manipulation datasets encompass real-robot teleoperation~\cite{fang2024rh20t,wu2025robomind,jiang2025galaxea}, human bimanual activity~\cite{kuehne2014language,zhan2024oakink2,carmona2025bicap}, and simulated interactions~\cite{james2020rlbench,mu2021maniskill,li2023behavior}.  In robot recordings that pair trajectories with video, kinematic signals provide temporal cues that complement the visual information used to interpret actions.  Together, these datasets offer diverse tasks and interaction patterns for evaluating the temporal and semantic quality of automatic labelling across manipulation settings.

\subsection{Temporal Localisation and Semantic Annotation}
Recent annotators use pretrained VL models to transfer language to demonstrations or to associate visual skills with keyframes~\cite{xiao2023dial,kuramshin2025task,kou2025roboannotatorx}.  A separate temporal-segmentation method learns action boundaries from video through multi-stage models, query-based architectures, or annotation interfaces~\cite{farha2019ms,wang2026timestamp,ding2023temporal,stanovcic2026atlas}.  The former offers semantic flexibility but leaves timestamp discovery to a model that is not designed for precise long-range localisation, and the latter offers temporal structure but typically relies on task-specific visual training or annotation.  Structured annotation formats are also discussed to show how actions, acting arms, objects, and state changes can be represented in a theoretical way~\cite{lu2026ssc}.   Our pipeline links motion-based localisation to semantic capability: trajectory heuristics supply candidate phases, and VL reasoning describes their contents. 

\section{DEMONSTRATION MODELLING}
\label{sec:formalism}


We focus here on the automatic procedure that instantiates those subtasks from raw demonstrations. By adopting the structured-record format introduced in the structured subtasks chain (SSC) formulation~\cite{lu2026ssc}, we represent a demonstration as
\begin{equation}
  \mathcal{D}=\bigl(\mathcal{X},\mathcal{V},\mathcal{O}\bigr),\;
  \mathcal{X}=\{\mathbf{x}_t\}_{t=0}^{T-1},
\end{equation}
where $\mathcal{X}$ contains synchronised per-arm end-effector poses, gripper commands, chassis pose, torso and arm-joint signals, $\mathcal{V}$ contains the available camera observations, and $\mathcal{O}$ is a task-relevant object inventory used to constrain object naming for simplicity. The construction maintains three physical streams
\begin{equation}
 \mathcal{Q}=\{\mathtt{base},\mathtt{left},\mathtt{right}\},\; \mathcal{A}=\{\mathtt{left},\mathtt{right}\},
\end{equation}
where the base and the two arms remain separate and their activity can overlap asynchronously.  A semantic stream $\mathcal{R}^q$ is represented by
\begin{equation}
 \mathcal{R}^q = \{\rho_j^q\}_{j=1}^{J},\; \rho_j^q=([u_j^q,v_j^q),z_j^q),\; q\in\mathcal{Q},
 \label{eq:stream-row}
\end{equation}
where $\rho_j^q$ is a tuple representing temporal interval $[u_j^q,v_j^q)$ and semantic content $z_j^q$, which contains a canonical action class, verb, target object, event type, and description. 
For an arm stream, the canonical class distinguishes approach, grasp, hold, release, retract, contact, tool use, and idle behaviour. The event field records an acquisition or release when supported by the gripper signal and visual evidence.  
For the base stream, the same record carries a navigation verb with a direction, targeting object or destination. 

This common schema keeps the temporal support and semantic fields aligned while allowing the three streams to evolve independently. 
Within-stream aggregation produces the final ordered stream collection $\overline{\mathcal{R}}=(\overline{\mathcal{R}}^q)_{q\in\mathcal{Q}}$. 

Overall, the automatic procedure follows the division of labour between physical and visual evidence:
\begin{equation}
\begin{aligned}
\mathcal{D}
&\xrightarrow{\text{trajectory segmentation}}
\mathcal{P}, \\
\mathcal{P}
&\xrightarrow{\text{phase-localised VL}}
\mathcal{R}=(\mathcal{R}^{\mathtt{base}},\mathcal{R}^{\mathtt{left}},\mathcal{R}^{\mathtt{right}}), \\
\mathcal{R}
&\xrightarrow{\text{within-stream aggregation}}
\overline{\mathcal{R}}.
\end{aligned}
\label{eq:construction-path}
\end{equation}
The segmentation stage $\mathcal{P}$ determines the temporal phases; the VL stage supplies semantic content inside each phase; and aggregation removes redundant cuts when adjacent descriptions in the same stream describe one continuing action. The realised framework is introduced in Section~\ref{sec:method}.

\section{AUTOMATIC ANNOTATION PIPELINE}
\label{sec:method}

\subsection{Kinematic Trajectory Segmentation}
\label{sec:segmentation}
Kinematic heuristics are used here to fastly locate physical changes in time.  The segmenter  analyses arm, base, torso, and gripper signals before any VL model is called for semantics.  Its stages are (i) per-signal segmentation and activity classification, (ii) cross-stream combination, and (iii) duration-aware shaping.  The result is the phase sequence $\mathcal{P}$ supplied to the VL reasoner.  

\subsubsection{Per-Signal Segmentation}
\label{sec:per-signal}
\begin{figure}[t]
\vspace{0.15cm}
  \centering
  \includegraphics[width=\columnwidth]{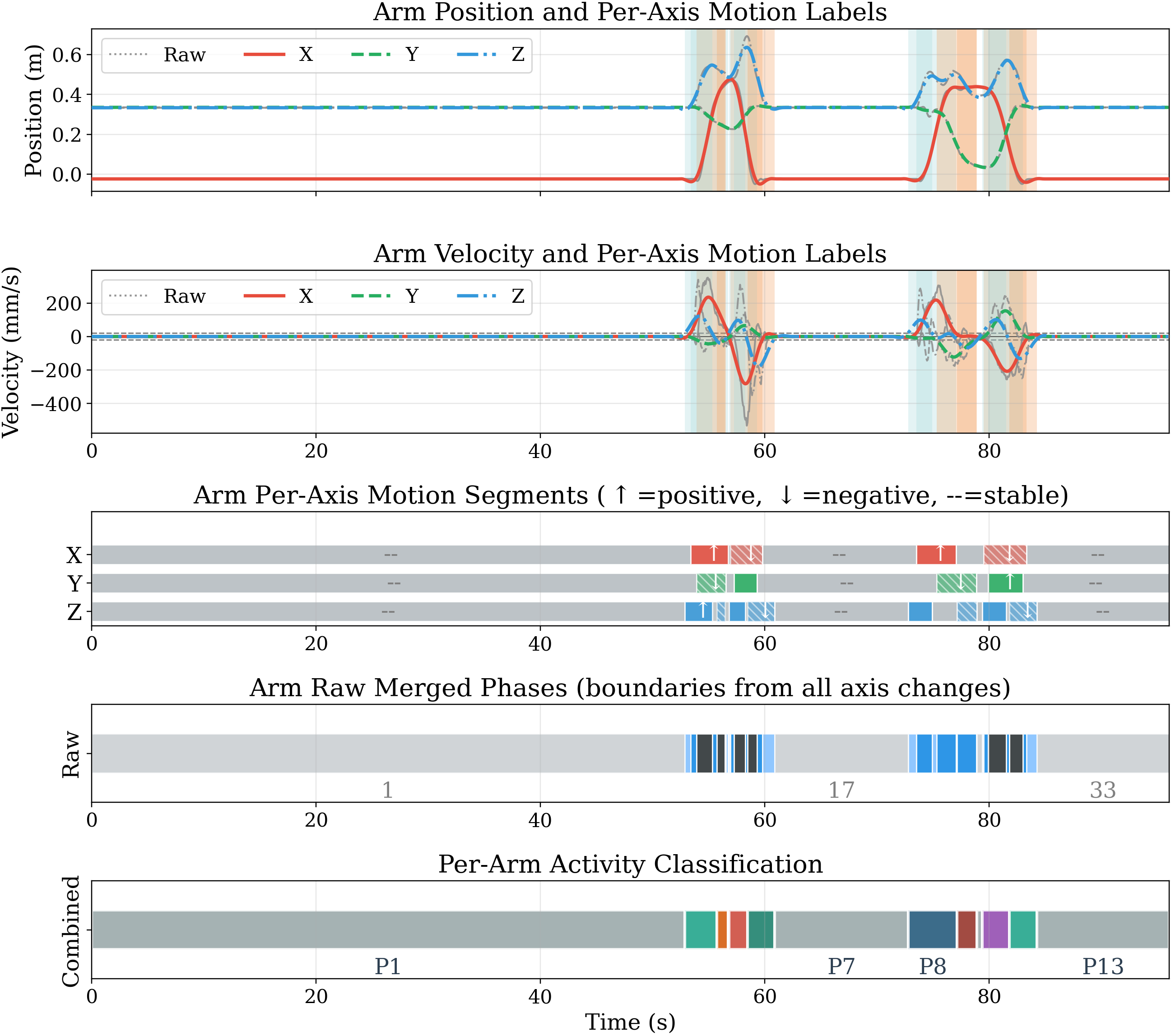}
  \caption{Direction-aware Cartesian arm segmentation for the Galaxea \emph{Doll Storage} task.  Position signals are first smoothed and differentiated (row 1). Then signed velocity thresholds label positive, negative, and stable motion on the three Cartesian axes (row 2-3).  Per-axis boundaries are unified into multi-axis segments (row 4), and compatible interruptions are bridged to form the final arm segments (row 5).}
  \label{fig:arm-segmentation}
\vspace{-0.5cm}
\end{figure}

The two arm end-effectors and the mobile base share a signed segmentation procedure following~\cite{lu2026keypose}. For an arm, $\mathbf{p}_t=(x_t,y_t,z_t)^{\mathsf T}$ in the robot-base frame; for the base, $\mathbf{p}_t=(x_t,y_t,\psi_t)^{\mathsf T}$, where $\psi_t$ is unwrapped yaw motion.  Each coordinate is first smoothed with a Savitzky--Golay filter, differentiated by a backward finite difference, and passed through a moving average:
\begin{equation}
 \begin{aligned}
 \widetilde{\mathbf p}_t&=\operatorname{SavGol}_t(\mathbf p;w_{\rm SG},r_{\rm SG}),
 &\mathbf v_0&=\mathbf 0,\\[-2pt]
 \mathbf v_t&=\frac{\widetilde{\mathbf p}_t-\widetilde{\mathbf p}_{t-1}}{\Delta t},
 &\overline{\mathbf v}_t&=\operatorname{MA}_t(\mathbf v;w_{\rm MA}).
 \end{aligned}
 \label{eq:velocity-filter}
\end{equation}
For coordinate $j$, the signed label is
\begin{equation}
 \ell_t^j=\operatorname{sign}(\overline v_t^j)\,\mathbf 1[|\overline v_t^j|>\theta^j],
 \; \ell_t^j\in\{-1,0,+1\}.
 \label{eq:signed-motion}
\end{equation}
As shown in Fig.~\ref{fig:arm-segmentation},  constant-sign runs are combined across the coordinates into direction-aware motion segments.  Short interruptions are bridged when the flanking segments share a same-sign active coordinate and contain no conflicting coordinate.  The per-axis sign signature is retained, since an immediate reversal by one arm can represent different actions.

\begin{figure}[t]
\vspace{0.15cm}
  \centering
  \includegraphics[width=\columnwidth]{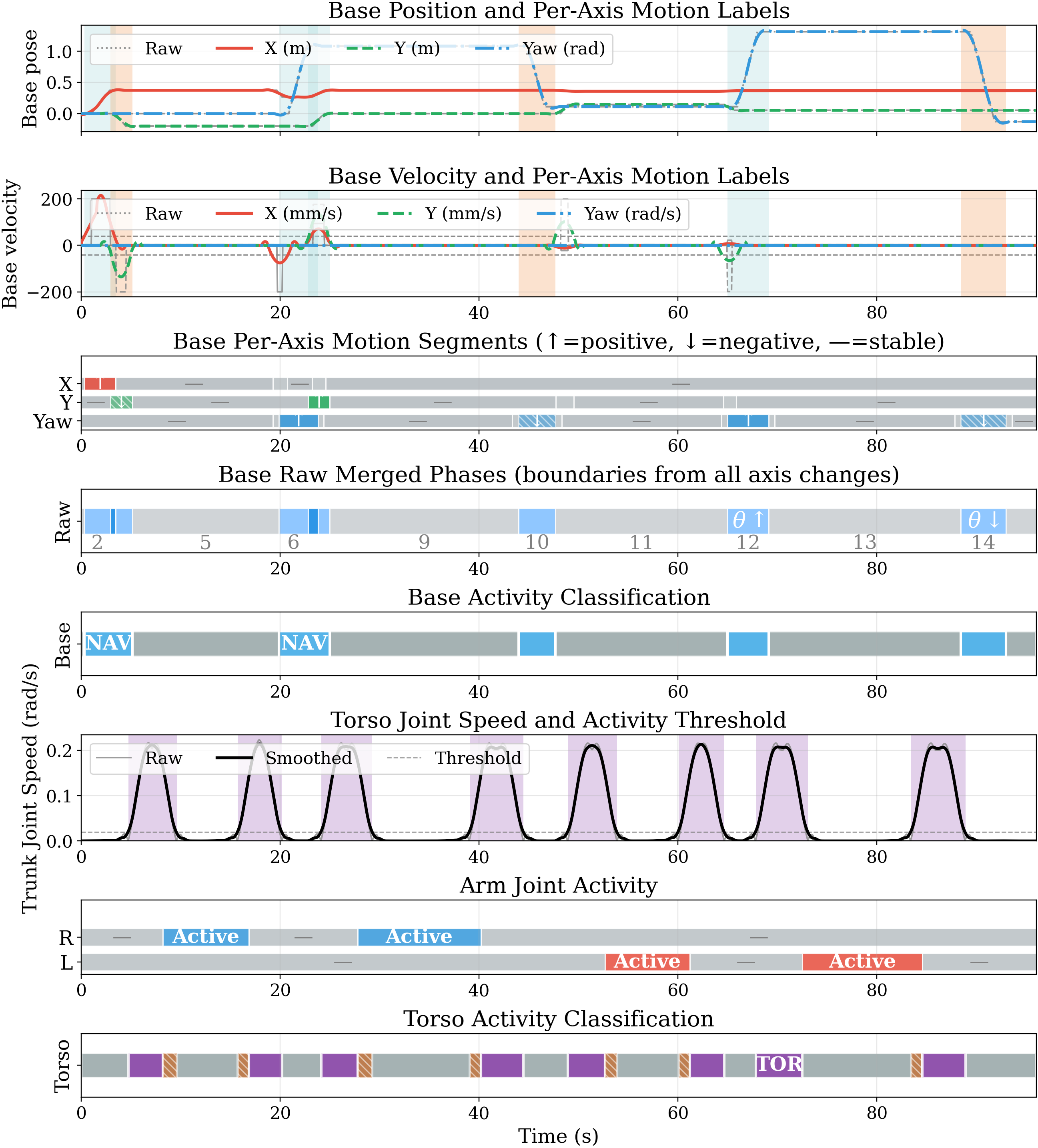}
  \caption{Base and torso segmentation for the Galaxea \emph{Doll Storage} task.  Base translation and yaw pass through signed segmentation (row 1-3). Then segments are merged and labelled (row 4-5).  Separately, torso speed is compared with the displayed right- and left-arm joint activity to identify torso-only segments (row 6-8).  The resulting $\mathsf{NAV}$ and $\mathsf{TORSO}$ segments are used in the subsequent cross-stream combination.}
  \label{fig:platform-torso}
\vspace{-0.5cm}
\end{figure}

Illustrated in Fig.~\ref{fig:platform-torso}, applying the same procedure to the base yields candidate base-motion segments labelled $\mathsf{NAV}$ from translational and yaw coordinates.  Meanwile, torso-active segments are labelled $\mathsf{TORSO}$ when joints of two arms have no activities, separating torso motion from arm motion.

\begin{figure*}[t]
\vspace{0.15cm}
  \centering
  \includegraphics[width=0.95\textwidth]{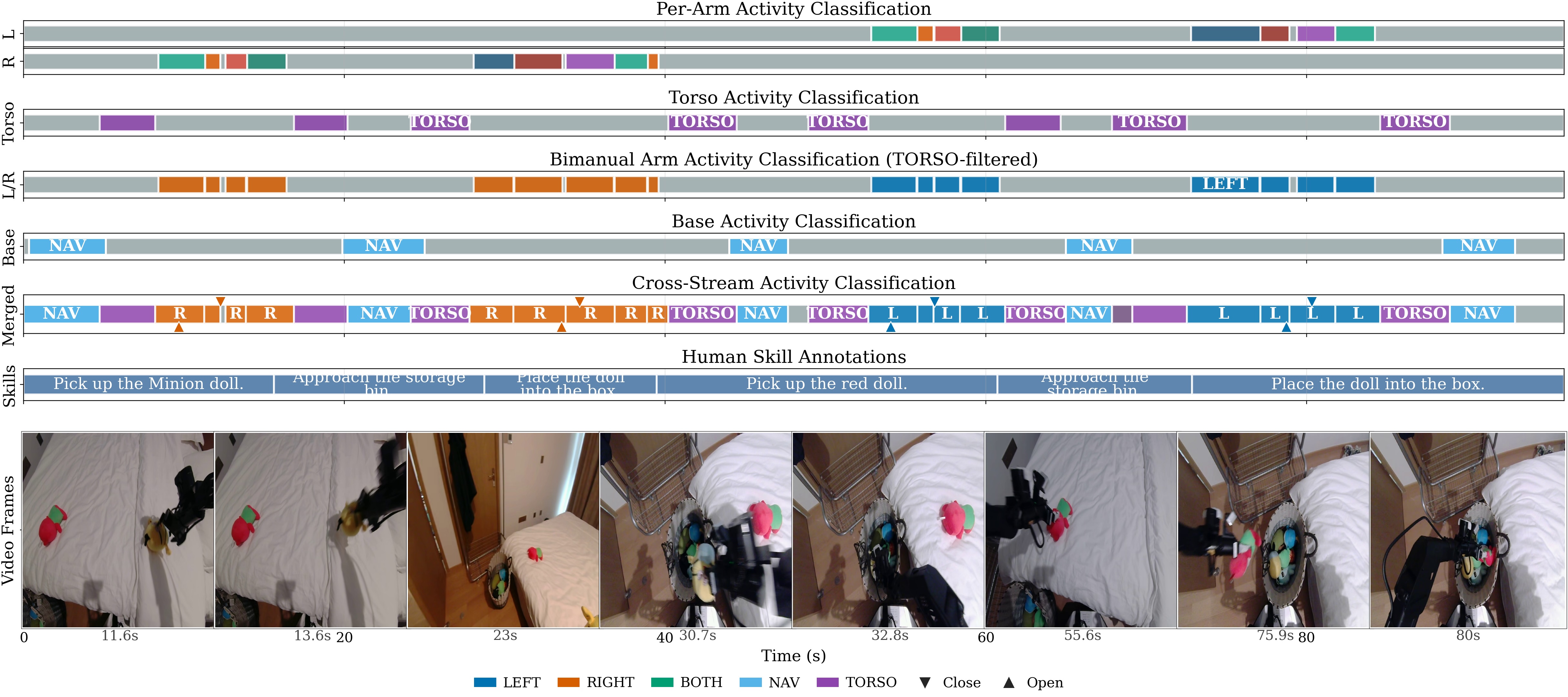}
  \caption{Cross-stream temporal construction for the Galaxea \emph{Doll Storage} task.  The left- and right-arm segments (row 1-2) are first filtered by the torso-only classification (row 3), then combined with the base (row 4) and torso segments under the activity-priority rule (row 5).  Concurrent lower-priority activity is retained as context.  Gripper events are associated with their containing phases, and duration-aware refinement produces the activity phases supplied to VL reasoning.  The source visualisation includes dataset annotations (row 6) and extracted video frames (row 7) for qualitative comparison.}
  \label{fig:cross-stream}
\vspace{-0.2cm}
\end{figure*}

\begin{figure*}[t]
  \centering
  \includegraphics[width=0.92\textwidth]{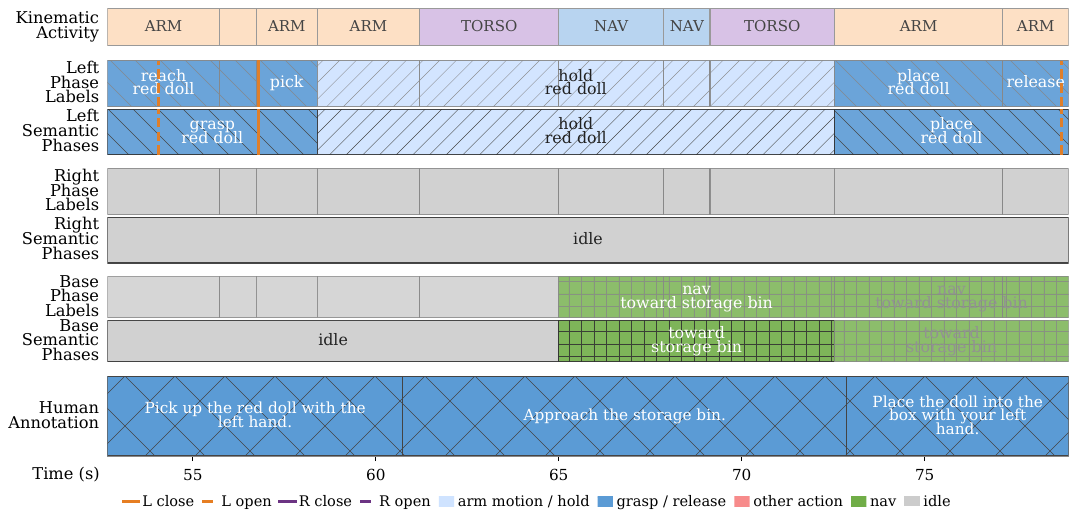}
  \caption{Phase-localised VL description and within-stream semantic aggregation for the Galaxea \emph{Doll Storage} excerpt.  Kinematic activity phases are feeded into VL calls (row 1); raw left-arm, right-arm, and base descriptions are then aggregated independently (row 2, 4, 6).  Aggregation merge the individual descriptions into a unified semantic representation (row 3, 5, 7).  The reference rail is qualitative context rather than a construction input (row 8).}
  \label{fig:phase-reasoning}
\vspace{-0.4cm}
\end{figure*}

\subsubsection{Cross-Stream Combination}
\label{sec:cross-stream}
The independently segmented streams have different boundaries and may overlap.  Cross-stream combination therefore produces a unified set of activity phases in three steps, as presented in Fig.~\ref{fig:cross-stream}.

\paragraph{Arm-state construction}
Before combination, $\mathsf{TORSO}$ segments are also used to filter corresponding arm segments so that torso-induced end-effector motion is represented only by the torso stream.
The resulting torso-filtered arm segments are combined as a pair
\begin{equation}
 \mathbf{b}_t=(b_t^{\mathtt{left}},b_t^{\mathtt{right}})\in\{0,1\}^2,
 \label{eq:bimanual-state}
\end{equation}
where $b_t^a$ indicates filtered activity of arm $a\in \mathcal{A}$.  The four combinations encode the categorical states $\mathsf{LEFT}$, $\mathsf{RIGHT}$, $\mathsf{BOTH}$, and $\mathsf{STABLE}$. Simultaneous activity is therefore retained as $(1,1)$ while the later semantic stages continue to treat the two arms as separate streams.  

\paragraph{Synchronous merge}
The merge of four streams leads to a synchronous phase sequence with categorical arm, base, and torso labels.
Let $\operatorname{arm}_t$ be the categorical arm state encoded by $\mathbf b_t$, and let $\operatorname{base}_t\in\{\mathsf{NAV},\mathsf{STABLE}\}$ and $\operatorname{torso}_t\in\{\mathsf{TORSO},\mathsf{STABLE}\}$.  At each frame, the priority order
\begin{equation}
 \mathsf{ARM}\succ\mathsf{TORSO}\succ\mathsf{NAV}\succ\mathsf{STABLE},
 \label{eq:activity-priority}
\end{equation}
selects the primary activity label.  Any concurrent lower-priority activity is retained within a \emph{context label}, such as base--arm coactivity or torso motion during arm activity.  Each maximal constant run of the primary label yields one initial activity phase, so the resulting phases are mutually exclusive and cover $[0,T)$.  Gripper sign changes provide one-frame close/open events that are attached to the phase containing their start frame.

\paragraph{Duration-aware shaping}
The initial phases can be too short (below time threshold) for an independent semantic query, so we further apply shaping rules as postprocessing. 
First, consecutive short phases without gripper events are clustered inheriting the highest-priority activity in Eq.~\ref{eq:activity-priority}.
Second, remaining singleton phases are absorbed into a neighbouring phase using the same priority order. 
The output $\mathcal{P}$ is dense, event-preserving, and suitable for phase-local reasoning.

\subsection{Phase-Localised Vision-Language Reasoning}
\label{sec:vl-reasoning}
Kinematic segmentation establishes each activity phase candidate before any semantic call, but cannot determine contents from motion alone.  Next, the VL model is applied to process phases chronologically to describe the contents.

For activity phase $P_i$, the response is represented following~\cite{lu2026ssc} as
\begin{equation}
  r_i=\bigl(d_i,m_i,\{a_i\}_{a\in\mathcal{A}},g_i\bigr),
 \label{eq:phase-response}
\end{equation}
where $d_i$ is a one-sentence visual summary, $m_i$ is an optional navigation target, and $a_i$ is an arm descriptor. Each arm descriptor contains a verb, target object, contact flag, and event type in $\{\texttt{ACQUIRE},\texttt{RELEASE},\texttt{OTHERS},\texttt{STABLE}\}$.  The model is not encouraged to name a special verb before the corresponding physical interaction is visible.
$g_i = (\,\mathrm{left}(g),\; \mathrm{right}(g),\; \mathrm{same}(g)\,)$ is a simplified scene graph to identify current state, with $\mathrm{left}(g), \mathrm{right}(g) \in \mathcal{O}$ represeting object identifiers held by each hand and $\mathrm{same}(g) \in \{\mathsf{T}, \mathsf{F}\}$ recording whether two hands hold the \emph{same} physical object.

The prompt supplies the fixed phase, $({\texttt{arm}},{\texttt{base}},{\texttt{torso}})$ labels, context labels, gripper events, and a motion summary when navigation is present.  It also supplies the previous description $r_{i-1}$ for consistency reference. 
For visual infomation, duration-adaptive sampling is adopted in seconds and renders each time point as the triptych views [left wrist $|$ head $|$ right wrist], combining local arm action views with head scene context.  
Stable phases are queried with an instruction to continue the preceding description unless visual evidence indicates a change.

\subsection{Within-Stream Semantic Aggregation}
\label{sec:row-merge}
After VL reasoning, we merge the three streams independently based on semantic aspects to produce the final ordered stream collection $\overline{\mathcal{R}}=(\overline{\mathcal{R}}^q)_{q\in\mathcal{Q}}$.  

Each arm phase description is normalised to one of the semantic classes $\mathsf{approach}$, $\mathsf{grasp}$, $\mathsf{hold}$, $\mathsf{release}$, $\mathsf{retract}$, $\mathsf{contact}$, $\mathsf{tool\mbox{-}use}$, or $\mathsf{idle}$. Each class can contain multiple free-form verbs, for example, $\mathsf{approach}$ includes \emph{reach}, \emph{hover}, \emph{move-to}, and \emph{extend}; and $\mathsf{retract}$ includes \emph{withdraw}, \emph{rest}, \emph{back-off}, and \emph{pull-back}.
Compatible descriptions are merged when their targets agree.  
To make action consistent, same-target \emph{approach} phases followed by an \emph{acquire} event are integrated into a $\mathsf{grasp}$ phase (in Fig.~\ref{fig:phase-reasoning}, row 3).  A $\mathsf{release}$ absorbs following arm $\mathsf{retract}$ phases.  
Meanwhile, the base stream independently merges compatible navigation phases and also represents the facing direction status of the robot.

Conservative rules are used to correct misinterpretations in case the VL model generates inconsistencies. First, an ambiguous \emph{press}, \emph{touch}, or \emph{adjust} is reinterpreted as $\mathsf{approach}$ or $\mathsf{retract}$ only when a contiguous acquire or release--retract anchor supports that interpretation. Second, a $\mathsf{place}$ description adjacent to $\mathsf{hold}$ is corrected to $\mathsf{hold}$ when without a $\mathsf{release}$ event. Third, a $\mathsf{release}$--$\mathsf{hold}$--$\mathsf{release}$ is merged when without an intervening gripper event, and $\mathsf{hold}$--$\mathsf{release}$--$\mathsf{hold}$ requires an arm-matched gripper opening event so the post-release $\mathsf{hold}$ can become $\mathsf{retract}$.

The resulting three chronological streams contain rows with frame ranges, verbs, canonical kinds, gripper references, and repair provenance.  These rows are the output evaluated in this paper.

\section{EVALUATION}
\label{sec:experiments}

The evaluation starts from 87 real bimanual mobile-manipulation demonstrations from the Galaxea Open-World Dataset~\cite{jiang2025galaxea}, comprising 29 tasks with three episodes per task.  The demonstrations are stored in LeRobot/RLDS format at 15\,Hz.  A dataset wrapper maps camera, arm, chassis, torso, and gripper fields into the common trajectory record used by our automatic annotation pipeline.

We run our pipeline modules, the trajectory segmenter, phase reasoner, and stream semantic aggregator, without native subtask labels.  All reported VL reasoning uses GPT-5.5 model.
We evaluate this pipeline through representative demonstrations, reproducibility diagnostics, and participant review of the generated semantic rows.

\subsection{Same-Input VL Stability}
\label{sec:eval-variance}
During VL reasoning, we select one episode from each of five representative Galaxea tasks covering different task types: 
\emph{Arrange the Fruits}, \emph{Boil the Water}, \emph{Clean Toilet}, \emph{Cooking Food in an Air Fryer}, and \emph{Doll Storage} to examine reasoning stability.
We repeat the reasoner three times with identical inputs and prompts.  Deterministic segmentation makes the phase list and extracted frames identical across repeats, isolating VL sampling noise.  For each phase, a field is counted as full agreement when all three outputs match, majority when two match, and split when all differ.  Results are shown in Table~\ref{tab:variance}.

The aggregate full-agreement rate is 87.4\% on Galaxea, with a 2.3\% hard three-way split rate.  Possession is identical in every phase, and event type agrees on 92.7--94.1\% of phases.  Variation is concentrated in free-form verbs, while possession state and event type are substantially more stable, which are used to aggregate adjacent phases within a stream.

\begin{table}[h]
\centering
  \caption{VL-output stability on Galaxea in three identical-input repeated runs.  Entries are full-agreement rates over reasoned phases.}
\label{tab:variance}
\small
\begin{tabular}{lc}
\toprule
Schema field & Galaxea \\
\midrule
base target & 98.6\% \\
left verb / object / event type & 63.0\% / 90.0\% / 94.1\% \\
right verb / object / event type & 69.9\% / 86.3\% / 92.7\% \\
left / right possession & 100.0\% / 100.0\% \\
\midrule
aggregate (full) & \textbf{87.4\%} \\
aggregate (hard split) & 2.3\% \\
\bottomrule
\end{tabular}
\end{table}

\begin{figure}[t]
\vspace{0.15cm}
  \centering
  \includegraphics[width=0.95\columnwidth]{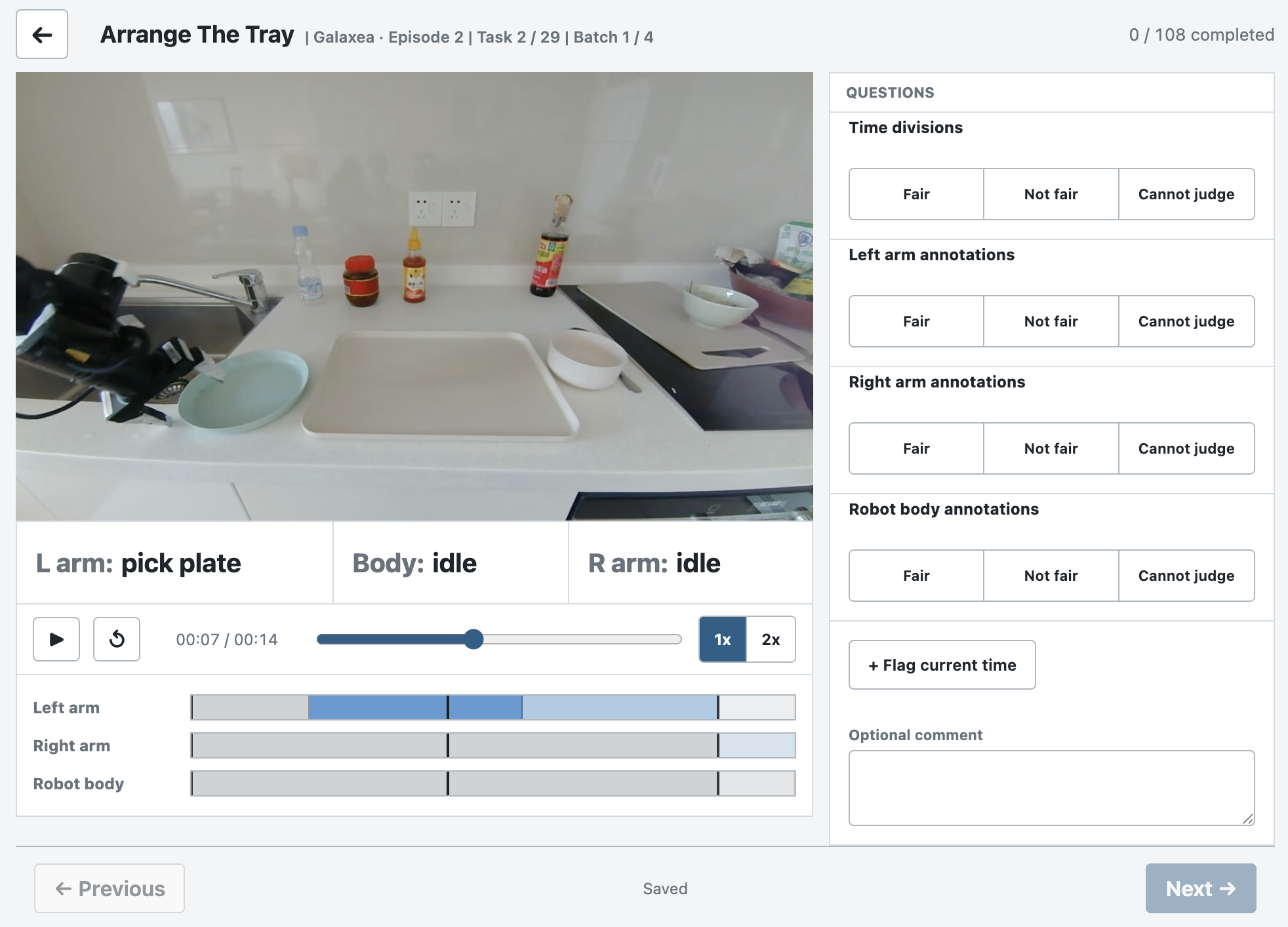}
  \caption{Participant review interface. The displayed video frame is reproduced without content alteration from the \protect\href{https://huggingface.co/datasets/OpenGalaxea/Galaxea-Open-World-Dataset}{Galaxea Open-World Dataset}~\cite{jiang2025galaxea} under \protect\href{https://creativecommons.org/licenses/by-nc-sa/4.0/}{CC BY-NC-SA 4.0}.}
  \label{fig:review-interface}
\vspace{-0.5cm}
\end{figure}

\subsection{Human Acceptability of Aggregated Semantic Rows}
\label{sec:eval-human}

\paragraph{Review design}
We propose four review questions to evaluate our pipeline in two aspects: temporal and semantic.  The temporal question asks whether the visible row boundaries divide the motion fairly, which probes quality of kinematic segmentation and phase shaping.  The left-arm, right-arm, and robot-body questions instead ask whether each active row correctly describes the action and 
target.  Responses were \texttt{Fair}, \texttt{Not fair}, or \texttt{Cannot judge}.

The review used the local web interface shown in Fig.~\ref{fig:review-interface} with the head-camera video, synchronised timelines and current descriptions for the two arms and robot body.  

The review covers episodes 0--2, the same 29-task catalogue per episode. The episode reviews contain 113, 117, and 108 demo clips. 
Nine participants were recruited, three for each episode.  Their self-reported profiles were heterogeneous: seven reported regular robotics experience and two reported none; annotation experience was reported as \emph{none} for six, \emph{some} for two, and \emph{regular} for one; and only one reported prior familiarity with Galaxea dataset.  The task order was shuffled per participant.

We report the judgeable fair rate $F/(F+N)$ and cannot-judge rate $C/(F+N+C)$.  For a semantic question, we exclude that question for a video clip when its corresponding arm or body stream is wholly idle; for the temporal question, we exclude a clip only when all three streams are wholly idle.  The retained rows and temporal divisions are called \emph{active}.  

\begin{table}[h]
\centering
\vspace{0.15cm}
\caption{Human review of generated rows and time divisions of the Galaxea dataset.  The judgeable fair rate is reported followed by raw counts $(F/N/C)$ for Fair, Not fair, and Cannot judge.}
\label{tab:human-acceptability}
\small
\renewcommand{\arraystretch}{0.93}
\begin{tabular}{clc}
\toprule
Episode & Review question & Fair rate $(F/N/C)$ \\
\midrule
0 & Robot body    & 90.7\% (98/10/0) \\
  & Left arm      & 75.6\% (167/54/1) \\
  & Right arm     & 80.6\% (212/51/1) \\
  & Time divisions& 92.4\% (304/25/7) \\
\addlinespace[1pt]
\midrule
1 & Robot body    & 93.1\% (95/7/0) \\
  & Left arm      & 79.3\% (161/42/1) \\
  & Right arm     & 74.5\% (181/62/3) \\
  & Time divisions& 88.5\% (308/40/3) \\
\addlinespace[1pt]
\midrule
2 & Robot body    & 87.7\% (71/10/0) \\
  & Left arm      & 80.1\% (161/40/0) \\
  & Right arm     & 82.1\% (192/42/0) \\
  & Time divisions& 90.7\% (291/30/0) \\
\midrule
All & Robot body    & 90.7\% (264/27/0) \\
    & Left arm      & 78.2\% (489/136/2) \\
    & Right arm     & 79.1\% (585/155/4) \\
    & Time divisions& 90.5\% (903/95/10) \\
\bottomrule
\end{tabular}
\end{table}

\begin{table}[t]
\centering
\vspace{0.15cm}
\caption{Reviewer consensus, reviewer background, and task-level results. Consensus items are applicable questions for one clip, each rated by three participants. The arm items combine both arms. Profile rates are equal-participant means, and task rates are equal-episode means.}
\label{tab:review-consensus}
\small
\setlength{\tabcolsep}{3.5pt}
\renewcommand{\arraystretch}{1.02}
\textit{(a) Reviewer consensus}\\
\begin{tabular}{@{}lccc@{}}
\toprule
Component & \shortstack{All three:\\Fair} & \shortstack{Split\\responses} & \shortstack{All three:\\Not fair} \\
\midrule
Time divisions (336) & 73.5\% & 26.2\% & 0.3\% \\
Robot-body labels (97) & 77.3\% & 21.6\% & 1.0\% \\
Arm labels (457) & 56.7\% & 38.9\% & 4.4\% \\
\bottomrule
\end{tabular}

\vspace{4pt}
\textit{(b) Reviewer familiarity}\\
\begin{tabular}{@{}llrcc@{}}
\toprule
Profile & Group & $n$ & Active semantics & Time divisions \\
\midrule
Annotation & Yes & 3 & 77.7\% & 86.3\% \\
Annotation & No  & 6 & 82.4\% & 92.5\% \\
Galaxea    & Yes & 1 & 81.8\% & 85.7\% \\
Galaxea    & No  & 8 & 80.7\% & 91.1\% \\
\bottomrule
\end{tabular}

\vspace{4pt}
\textit{(c) Representative task contrasts}\\
\begin{tabular}{@{}lcc@{}}
\toprule
Task & Temporal & Semantic \\
\midrule
Adjust air-conditioner temperature & 100.0\% & 66.7\% \\
Fold clothes & 96.2\% & 66.0\% \\
Floor-cloth wiping stains & 92.6\% & 92.8\% \\
Chair push and place & 75.6\% & 75.9\% \\
\bottomrule
\end{tabular}
\end{table}

\paragraph{Temporal localisation and semantic interpretation.}
Table~\ref{tab:human-acceptability} shows a consistent separation between temporal and arm-label acceptability: participants accept 90.5\% of judgeable time divisions, compared with 78.7\% of judgeable active arm labels. Robot-body labels approach the temporal result, and the relatively lower acceptance of arm labels recurs in every episode. This pattern supports the usefulness of the kinematic temporal structure while the interpretation of arm actions and their targets is more challenging. Excluding wholly idle streams (99.5\% and 100\% for the left and right arms) reduced interference‌ in describing active manipulation.

\paragraph{Reviewer consensus and scrutiny.}
Table~\ref{tab:review-consensus}(a) shows that arm labels receive less unanimous acceptance and more split responses than timing or body labels. Unanimous rejection is at low levels, so the acceptance gap also reflects differences in how reviewers assess the same descriptions. 
As shown in Table~\ref{tab:review-consensus}(b), participants with annotation experience gave lower average acceptance than those without it, especially for semantic labels. We believe that this is because of greater scrutiny based on professional expertise. 
These observations suggest that perceived semantic quality depends on both the interpretation of the demonstrated interaction and the criteria used to judge whether a description captures it adequately. 

\paragraph{Variation across tasks.}
The task comparisons in Table~\ref{tab:review-consensus}(d) distinguish difficulties in temporal organisation from those in semantic description. \emph{Folding clothes} and \emph{air-conditioner} adjustment have highly accepted timing but weaker semantic labels, showing that acceptable temporal divisions do not necessarily yield adequate descriptions of the interaction. \emph{Floor-cloth wiping} is well accepted on both dimensions, whereas \emph{chair pushing} perfoms good but with relatively low acceptance. 
Thus, the trajectory segmentation module can perform more trustable results, with semantic interpretation being more challenging. This motivates increasing robustness in both temporal and semantic dimensions to extend the pipeline to difficult tasks.

\begin{figure*}[t]
\vspace{0.15cm}
  \centering
  \includegraphics[width=0.96\textwidth]{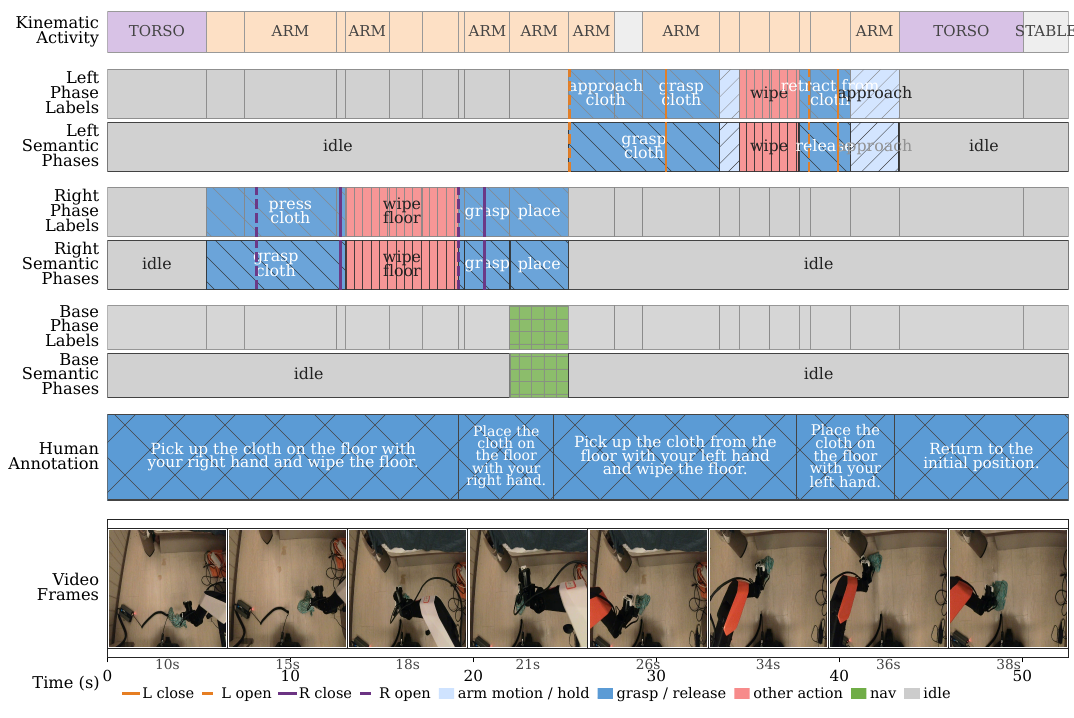}
  \caption{Phase-localised VL reasoning and within-stream semantic aggregation for Galaxea \emph{Floor Cloth Wiping Stains} task.  Independent left-arm, right-arm, and base streams retain their kinematic boundaries. The human annotation is contextual only.}
  \label{fig:galaxea-floor}
\vspace{-0.4cm}
\end{figure*}

The \emph{floor-cloth wiping} demonstration in Fig.~\ref{fig:galaxea-floor} illustrates the intended operating regime: fixed kinematic phases localise grasping, wiping, release, and repositioning, while independent arm and body streams preserve asynchronous motion.  
The same qualitative pattern appears in \emph{chair pushing} examples, where the body row records navigation separately from the arm's grasp--hold--release sequence.  
Deformable-object manipulation, such as \emph{arranging sofa cushions} and \emph{folding clothes}, is more difficult: persistent contact and small bimanual adjustments can produce several locally plausible descriptions that remain distinct after aggregation.

\begin{figure*}[t]
\vspace{0.15cm}
  \centering
  \includegraphics[width=0.96\textwidth]{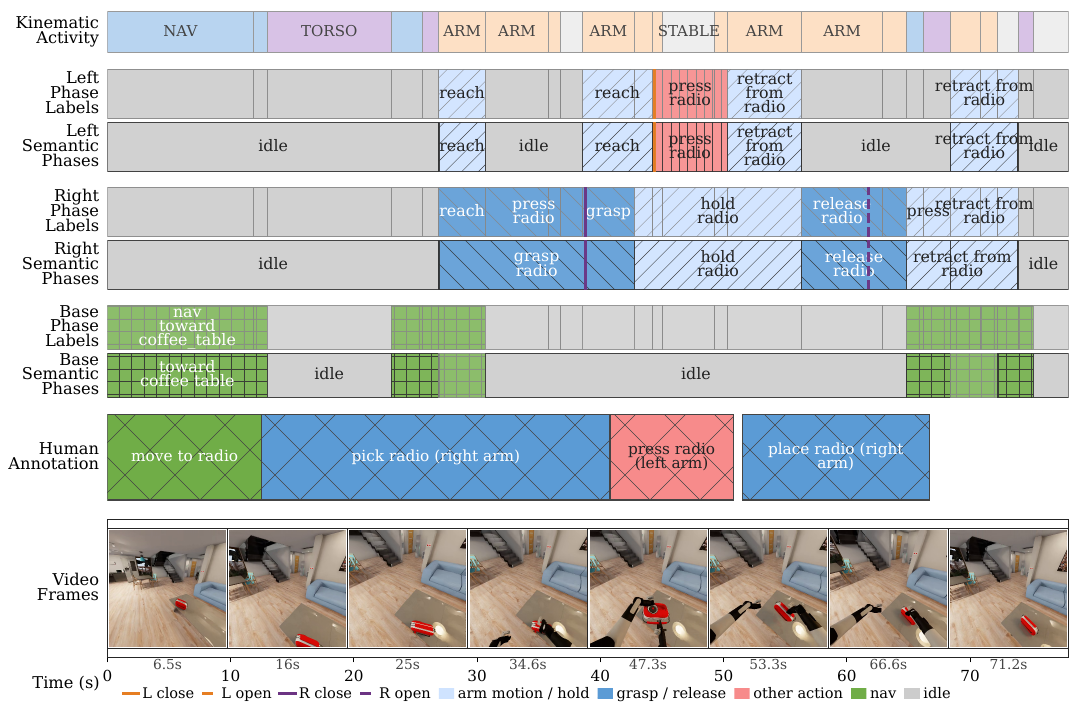}
  \caption{Portability example on BEHAVIOR-1K \emph{Turn On Radio} task. The same pipeline parameters are used here as in Galaxea dataset. The figure illustrates the potential reuse ability of the pipeline on different datasets.}
  \label{fig:b1k-portability}
\vspace{-0.4cm}
\end{figure*}

We also tested our pipeline on \emph{BEHAVIOR-1K} dataset for portability.  As shown in Fig.~\ref{fig:b1k-portability}, an episode was processed with the same parameters, implying the robostness of those three modules. Its 30\,Hz recording rate is converted to seconds for uniformity of duration parameters.

\section{LIMITATIONS AND FUTURE WORK}
\label{sec:limitations}

We have to admit that this ambitious pipeline is far from perfection. It is optimistic that this framework can benefit from the fast evolving VL model for more advanced performance. Still, many aspects are worth investigation, mainly in two directions: stronger automatic annotation and trustworthy verification.

The current pipeline now produces local arm and body descriptions, which do not provide a unified account of coordinated activity. Cross-stream composition is needed to assemble overlapping stream outputs into a coherent description of bimanual behaviour.

A second VL reasoning round could convert local descriptions outputed from our pipeline into a more general action sequence, reconcile repeated or synonymous labels, and verify apparent grasp events that lack supporting visual or state evidence. 

For visual input and prompt design, sparse frame sampling can miss brief or subtle interactions, and previous-phase visual and word context can propagate an unsupported interpretation. Action-adaptive sampling, evidence-aware look-back, prompt perturbations, and response-schema ablations could improve stability across tasks and repeated runs.

For tactile input, including tactile or force sensing will be greatly helpful to distinguish different types of contact or grasp actions, especially for deformable-object manipulation, which is the hardest part in segmentation and vl reasoning.  

Locally plausible descriptions may still imply inconsistent acquisition, release, or object possession across streams. Connecting the annotations to the state-based verification framework like SSC~\cite{lu2026ssc} would support cross-stream assembly, state propagation, ambiguity resolution, and consistency checking.

\section{CONCLUSION}
\label{sec:conclusion}

We presented an automatic labelling pipeline for bimanual mobile manipulation that combines kinematic temporal localisation with vision-language semantic interpretation. 
Trajectory segmentation establishes activity phases, phase-localised VL reasoning describes the observed interactions, and within-stream aggregation produces semantic annotations while preserving asynchronous arm and body behaviour. 
Evaluation on Galaxea shows stable possession and event fields across repeated runs and positive participant acceptance of temporal divisions, arm and body labels. This pipeline provides a structured basis for subsequent richer semantic subtask identification and state-based verification.

\section*{ACKNOWLEDGMENT}
This work involves Codex and Claude Code for both code and documentation preparation except bibliography. Careful line-by-line review and adjustments were made for clarity and accuracy for the whole contents in this paper.

\bibliographystyle{IEEEtran}
\bibliography{main}

\end{document}